\documentclass[runningheads]{llncs}
\usepackage[T1]{fontenc}
\usepackage{graphicx}
\usepackage{amsmath,amssymb}

\begin{document}

\title{CirrGuide: A Deep Cascaded Framework for Liver Cirrhosis Segmentation and Severity Classification from T2-Weighted MRI}

\titlerunning{CirrGuide: A Deep Cascaded Framework for Liver Cirrhosis Analysis}
\author{
Muntaqim Ahmed Raju\orcidID{0000-0002-2529-1233}\thanks{Corresponding author.}
\and
Ruizhe Ma\orcidID{0000-0003-2749-3063}
}


\authorrunning{M. A. Raju and R. Ma}

\institute{
University of Massachusetts Lowell, Lowell, MA 01854, USA\\
\email{\{MuntaqimAhmed\_Raju,Ruizhe\_Ma\}@uml.edu}
}

\maketitle              

\begin{abstract}
We present CirrGuide, a deep cascaded framework for cirrhotic liver segmentation and
severity classification. Cirrhosis causes
progressive structural changes in the liver and can lead to serious clinical
complications, making severity assessment important for disease monitoring and treatment
planning. However, severity classification is challenging because imaging patterns are
often subtle, spatially variable, and similar across adjacent stages. CirrGuide addresses
this by explicitly linking localization with classification. A ResNet50 encoder
with an Attention U-Net decoder first predicts a soft cirrhotic liver mask, which is then
used as an anatomical prior in a ResNet50-based classification branch. This branch
combines global multi-scale features with mask-guided attention-pooled regional features
to classify Mild, Moderate, and Severe cirrhosis. On the official CirrMRI600+ T2-weighted (T2W) 2D
split, CirrGuide achieves 89.83\% Dice and 84.14\% mIoU for segmentation, 69.58\%
accuracy and 61.55\% macro F1-score for severity classification. Compared with segmentation-only, classification-only,
and multi-task baselines, CirrGuide improves both localization and severity
classification, demonstrating the benefit of using predicted cirrhotic liver masks as
anatomical priors for cirrhosis analysis.

\end{abstract}


\section{Introduction}

Liver cirrhosis is a chronic and progressive liver disease that substantially affects
global health, causing approximately two million deaths each year and accounting for
nearly 2.4\% of global mortality~\cite{pellicoro2014liver,tsochatzis2014liver,zeng2025liver,huang2023global}. It represents the
end stage of many chronic liver disorders and can lead to serious complications such as
hepatic decompensation, liver cancer, and death~\cite{zeng2025liver}. Therefore, accurate assessment of
cirrhosis is important for detecting disease progression, supporting treatment planning,
and reducing the risk of life-threatening complications.

MRI is an important modality for liver assessment. It provides high soft-tissue contrast and enables precise, non-invasive evaluation of liver structure, supporting the diagnosis and staging of cirrhosis~\cite{ramalho2017magnetic,zeng2025liver}. In particular, T2-weighted (T2W) MRI can highlight tissue heterogeneity associated with fibrosis, edema, and fluid content, making it valuable for cirrhotic liver analysis~\cite{hshiao2012quantifying}.

Prior work on MRI-based fibrosis and cirrhosis staging has commonly used handcrafted radiomics or direct image-level classification. Radiomics- and CNN-based approaches using contrast-enhanced MRI have shown that fibrosis stage can be estimated from hepatobiliary-phase appearance~\cite{yasaka2018liver,park2019radiomics}. Radiomic features extracted from the liver and spleen have also been investigated for severity stratification in end-stage liver disease~\cite{nitsch2021mri}. However, image-level classification may include information from surrounding anatomical structures, whereas segmentation can help localize the tissue relevant to the prediction task. Nowak et al.~\cite{nowak2021detection} reported that liver segmentation before classification improved cirrhosis detection from standard T2W MRI by directing the model toward hepatic tissue. Gupta et al.~\cite{gupta2026advanced} also examined the use of liver localization for advanced fibrosis detection. These studies, however, mainly considered binary disease detection rather than multi-stage severity assessment, partly because of the limited availability of cirrhosis-specific MRI datasets.

The recently introduced CirrMRI600+ dataset supports cirrhotic liver analysis by providing T1-weighted (T1W) and T2W MRI scans with physician-refined liver masks and severity annotations~\cite{jha2025large}. Using this dataset, Zeng et al.~\cite{zeng2025liver} evaluated deep learning and radiomics-based models for cirrhosis staging. Their findings indicated that three-stage classification remains difficult, particularly for intermediate-stage disease, and that performance on T2W MRI was lower than on T1W MRI. Although this study provides a useful benchmark, the evaluated classifiers mainly operate at the image-level and do not use segmentation masks to guide feature extraction. Related studies have reported benefits from segmentation-guided learning for medical image classification by directing feature learning toward anatomically relevant regions~\cite{lin2023improving,rizhko2024improving,kumar2025improving,hossain2023thorax}.

Building on this, we propose CirrGuide, a cascaded segmentation-guided framework for cirrhotic liver region segmentation and severity classification from T2W MRI. CirrGuide first predicts a soft liver mask using a ResNet50 encoder and an Attention U-Net decoder. The predicted mask is then used as an anatomical prior in a ResNet50-based classification branch. Global multi-scale features are combined with mask-guided attention-pooled regional features, allowing the model to retain whole-image context while giving greater attention to features within the predicted liver region. In this way, CirrGuide links anatomical localization with multi-stage severity prediction.

Our main contributions are summarized as follows:
\begin{itemize}
    \item A cascaded segmentation-guided framework for cirrhotic liver region segmentation and severity classification from 2D T2W MRI.
    \item An anatomical-prior mechanism that uses the predicted soft cirrhotic liver mask to guide severity classification.
    \item Comprehensive evaluation of the framework on the official CirrMRI600+ 2D T2W MRI split against segmentation, classification, and multi-task baselines, with slice-level, patient-level, and qualitative analyses.
\end{itemize}

\section{Method}

\begin{figure}[!htbp]
\centering
\includegraphics[width=\textwidth]{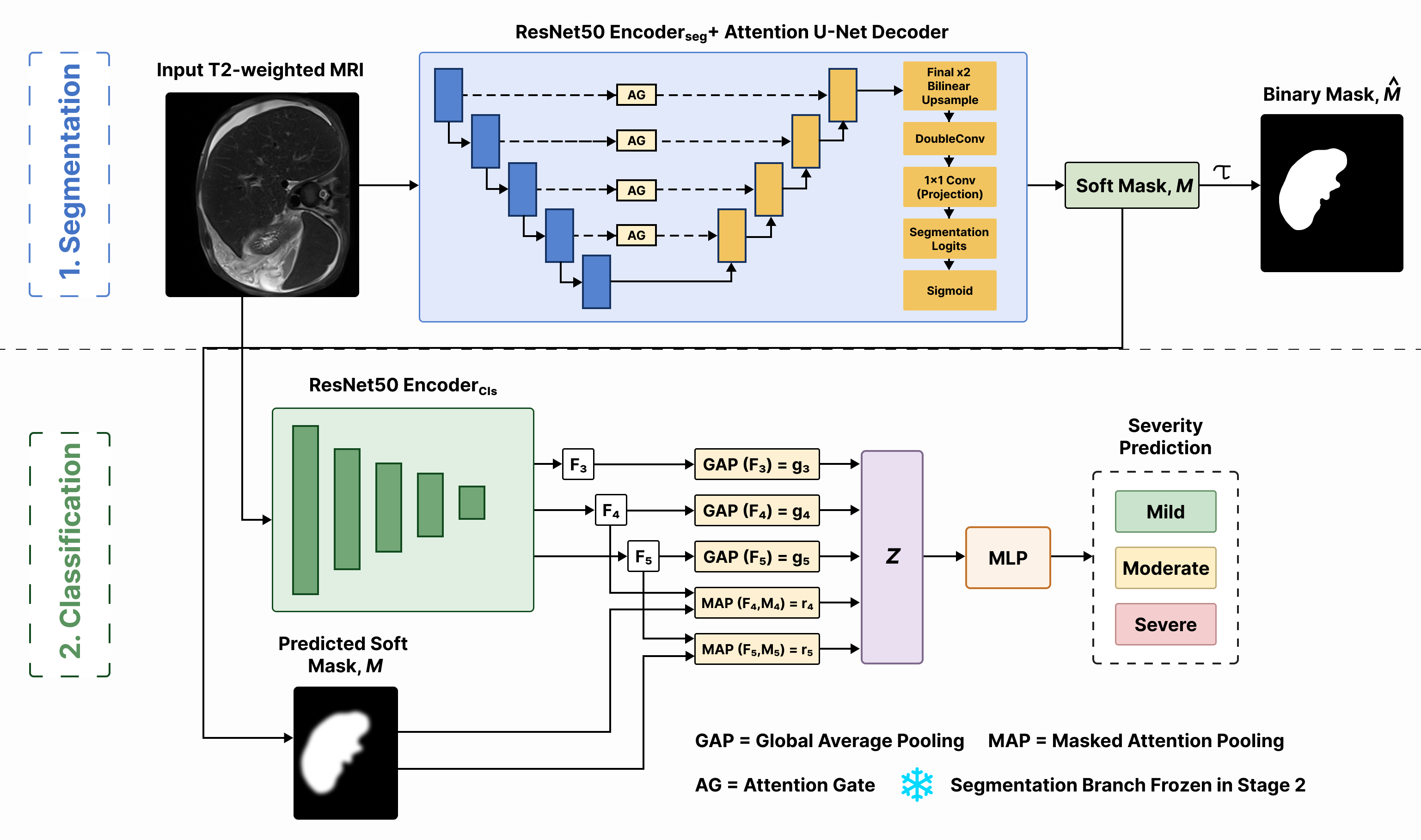}
\caption{Overview of CirrGuide. Stage~1 predicts the cirrhotic liver mask using a ResNet50 encoder and Attention U-Net decoder. In Stage~2, the frozen segmentation branch provides a soft mask prior for attention pooling over $F_4$ and $F_5$, while global features are extracted from $F_3$--$F_5$. The resulting global and regional features are fused for three-stage severity classification.}
\label{fig:framework}
\end{figure}

We propose \textbf{CirrGuide}, a deep cascaded framework that first segments the cirrhotic
liver region and then uses the predicted soft mask to guide severity classification
(Fig.~\ref{fig:framework}).

Given a T2W MRI slice $x$, the segmentation branch
predicts a soft cirrhotic liver mask:
\begin{equation}
M = \sigma(f_{\mathrm{seg}}(x)),
\end{equation}
where $f_{\mathrm{seg}}(\cdot)$ is implemented using a ResNet50 encoder with an
attention U-Net decoder, and $M \in [0,1]^{1 \times H \times W}$ denotes the
predicted soft mask. The final binary segmentation prediction is obtained as
\begin{equation}
\hat{M} = \mathbf{1}(M > \tau),
\end{equation}
where $\tau$ is the segmentation threshold.
The segmentation branch is trained with a combination of Dice loss
and binary cross-entropy loss:
\begin{equation}
\mathcal{L}_{\mathrm{seg}}
=
0.5\mathcal{L}_{\mathrm{Dice}}
+
0.5\mathcal{L}_{\mathrm{BCE}}.
\end{equation}

We assign equal weights to Dice loss and BCE loss so that region-level overlap and
pixel-wise supervision contribute equally to segmentation training. The BCE term uses a positive-pixel weight of 5.0 to reduce foreground-background imbalance. After segmentation training, the segmentation branch is frozen and used to provide
anatomical guidance for classification. A ResNet50 classification encoder extracts a
hierarchical feature pyramid:
\begin{equation}
\{F_1,F_2,F_3,F_4,F_5\}=f_{\mathrm{enc}}^{\mathrm{cls}}(x).
\end{equation}
We use $F_3$, $F_4$, and $F_5$ for severity prediction, as these deeper features balance
spatial resolution with semantic depth, whereas $F_1$ and $F_2$ mainly capture
low-level edges and local intensity patterns. Global contextual features are then
obtained by average pooling:
\begin{equation}
g_k = \mathrm{GAP}(F_k), \quad k \in \{3,4,5\}.
\end{equation}

To focus classification on anatomically relevant areas, CirrGuide
applies mask-guided attention pooling to $F_4$ and $F_5$. For each feature level $k \in
\{4,5\}$, the predicted mask $M$ is resized to the spatial resolution of $F_k$ to obtain
$M_k$. A lightweight attention module computes feature-dependent spatial logits, which
are combined with the soft mask prior:
\begin{equation}
\tilde{A}_k
=
\sigma\left(
W_k^{(2)}\delta(W_k^{(1)}F_k) + M_k
\right),
\end{equation}
where $W_k^{(1)}$ and $W_k^{(2)}$ are $1\times1$ convolutional layers and
$\delta(\cdot)$ denotes ReLU activation. The segmentation-guided regional descriptor is
then computed as:
\begin{equation}
r_k =
\frac{
\sum_{h,w} F_k(:,h,w)\tilde{A}_k(h,w)
}{
\sum_{h,w}\tilde{A}_k(h,w)+\epsilon
}.
\end{equation}
This operation uses the predicted cirrhotic liver mask as a soft spatial prior while
allowing the classifier to learn non-uniform attention over severity-relevant
subregions. The final classification representation combines global image context and mask-guided
regional features:
\begin{equation}
z = [g_3 \Vert g_4 \Vert g_5 \Vert r_4 \Vert r_5],
\end{equation}
where $\Vert$ denotes concatenation. The fused feature vector is passed to an MLP
classifier:
\begin{equation}
\hat{p}=\mathrm{softmax}(f_{\mathrm{cls}}(z)).
\end{equation}

The classification branch predicts three severity classes: Mild, Moderate, and Severe.
It is trained using cross-entropy:
\begin{equation}
\mathcal{L}_{\mathrm{cls}}
=
-\sum_{c=1}^{3} y_c\log(\hat{p}_c),
\end{equation}
where $y_c$ denotes the one-hot severity label. Weighted sampling is used during
training to reduce class imbalance.

\section{Experimental Evaluation}

\subsection{Data}

We evaluate CirrGuide on the official T2W 2D split of CirrMRI600+, which
provides abdominal MRI slices with physician-refined cirrhotic liver masks and severity
labels. The official case-level split contains 256 training, 31 validation, and 31
testing cases, corresponding to 5,428, 674, and 664 2D slices, respectively. At the case
level, the Mild, Moderate, and Severe counts are 102, 99, and 55 for training; 13, 9,
and 9 for validation; and 16, 6, and 9 for testing. At the slice level, the
corresponding counts are 2,215, 2,097, and 1,116 for training; 309, 189, and 176 for
validation; and 350, 133, and 181 for testing.

\subsection{Implementation Details}

\textbf{Inputs:} All T2W MRI slices and masks are cropped, resized to $256
\times 256$, normalized, and augmented during training using alignment-preserving
transformations. \textbf{Training:} CirrGuide is trained in a cascaded manner. The
segmentation branch is first trained to predict the cirrhotic liver mask. After
training, the segmentation branch is frozen and used to generate soft masks for the
classification branch. The classification branch then uses these predicted masks as
anatomical priors for mask-guided attention pooling. Classification-only baselines are
trained using the same image inputs without segmentation guidance, while multi-task
baselines are trained by jointly optimizing segmentation and classification objectives.
\textbf{Hyper-parameters:} All implemented models were trained for up to 40 epochs with early stopping, using a batch size of 12, learning rate of $1\times10^{-4}$, and weight decay of $1\times10^{-4}$. We
use the AdamW optimizer and automatic mixed precision training. Weighted sampling is
used for classification and multi-task training to reduce class imbalance. The best
classification model is selected based on validation macro F1-score. All experiments are
implemented in PyTorch and trained on a single NVIDIA RTX A6000 GPU with a fixed random
seed of 42. \textbf{Evaluation:} Segmentation
performance is evaluated using Dice score and mean intersection over union (mIoU).
Severity classification is evaluated using accuracy, macro-averaged precision, recall,
and F1-score to account for class imbalance. Patient-level performance is computed by
averaging slice-level softmax probabilities across all slices from the same test case
and assigning the final label using the highest averaged probability.

\subsection{Results and Discussion}

\textbf{Segmentation Performance:} Table~\ref{tab:seg_results} reports cirrhotic liver
region segmentation performance on the official CirrMRI600+ T2W 2D test split.

\begin{table}[!htbp]
\centering
\caption{Segmentation performance on the CirrMRI600+ T2W 2D test split.}
\label{tab:seg_results}
\setlength{\tabcolsep}{8pt}
\begin{tabular}{lcc}
\hline
\textbf{Method} & \textbf{Dice} & \textbf{mIoU} \\
\hline
MedSegDiff~\cite{jha2025large} & 0.7667 & 0.7489 \\
Attention U-Net & 0.8626 & 0.7936 \\
U-Net & 0.8967 & 0.8367 \\
ResNet50 TransUNet & 0.8911 & 0.8286 \\
\hline
Multi-task U-Net & 0.8343 & 0.7540 \\
Multi-task ResNet50 TransUNet & 0.8790 & 0.8093 \\
Multi-task Attention U-Net & 0.8224 & 0.7403 \\
\hline
\textbf{CirrGuide Segmentation} & \textbf{0.8983} & \textbf{0.8414} \\
\hline
\end{tabular}
\end{table}

CirrGuide achieved the best segmentation performance, with a Dice score of 0.8983 and an mIoU of 0.8414. Compared with MedSegDiff, the strongest 2D benchmark reported by Jha et al.~\cite{jha2025large}, it improved the Dice score by 17.17\% and the mIoU by 12.35\%. Among the segmentation-only baselines, U-Net and ResNet50 TransUNet also demonstrated strong performance, highlighting the effectiveness of encoder–decoder architectures for cirrhotic liver localization. The strongest multi-task baseline, Multi-task ResNet50 TransUNet, achieved a Dice score of 0.8790 and an mIoU of 0.8093; CirrGuide improved these metrics by 2.20\% and 3.97\%, respectively. The lower segmentation performance of the multi-task variants compared with their segmentation-only counterparts suggests that joint optimization introduces a trade-off between localization and severity classification. The proposed CirrGuide segmentation branch achieved more accurate localization of cirrhotic regions, which is particularly important because the resulting soft mask subsequently serves as an anatomical prior for severity classification.

\textbf{Slice-level Severity Classification:} Table~\ref{tab:cls_results} compares
CirrGuide with prior work, classification baselines, and multi-task baselines. CirrGuide
achieved the best slice-level result, with 0.6958 accuracy and 0.6155 macro F1-score.

\begin{table}[!htbp]
\centering
\caption{Slice-level severity classification performance on the CirrMRI600+ T2W 2D test set. Except for accuracy, all reported metrics are macro-averaged.}
\label{tab:cls_results}
\setlength{\tabcolsep}{6pt}
\begin{tabular}{lcccc}
\hline
\textbf{Method} & \textbf{Acc.} & \textbf{Prec.} & \textbf{Rec.} & \textbf{F1} \\
\hline
MambaVision-T~\cite{zeng2025liver} & 0.6380 & 0.5913 & 0.5801 & 0.5843 \\
\hline
DenseNet121 & 0.6039 & 0.5236 & 0.5286 & 0.5219 \\
U-Net classifier & 0.6310 & 0.4972 & 0.5357 & 0.5127 \\
ResNet50 & 0.5979 & 0.5752 & 0.5570 & 0.5577 \\
\hline
Multi-task U-Net & 0.5708 & 0.5866 & 0.5518 & 0.5498 \\
Multi-task ResNet50 TransUNet & 0.6024 & 0.6044 & 0.5624 & 0.5670 \\
Multi-task Attention U-Net & 0.6054 & 0.5476 & 0.5428 & 0.5414 \\
\hline
\textbf{CirrGuide Classification} & \textbf{0.6958} & \textbf{0.6202} & \textbf{0.6120} & \textbf{0.6155} \\
\hline
\end{tabular}
\end{table}

Compared with the MambaVision-T results reported by Zeng et al.~\cite{zeng2025liver}, CirrGuide achieved higher accuracy (0.6958 vs. 0.6380) and macro F1-score (0.6155 vs. 0.5843). As the evaluation splits differ, these results are presented as a reference comparison rather than a direct head-to-head evaluation. Compared with ResNet50, the strongest classification baseline in macro F1-score, CirrGuide improved accuracy from 0.5979 to 0.6958 and macro F1-score from 0.5577 to 0.6155, corresponding to relative gains of 16.37\% and 10.36\%, respectively. It also outperformed the strongest multi-task baseline, ResNet50 TransUNet, by 15.50\% in accuracy and 8.55\% in macro F1-score. These gains highlight the benefit of explicitly using the predicted soft mask as an anatomical prior for mask-guided attention pooling. In contrast, multi-task baselines jointly optimize segmentation and classification, which may introduce competing objectives and require a trade-off between the two tasks. Moreover, they do not directly use the predicted mask to guide classification. CirrGuide instead preserves global context while emphasizing liver regions most relevant to severity prediction.

Table~\ref{tab:classwise_results} presents the class-wise slice-level performance. Performance varied substantially across severity stages. CirrGuide achieved strong results for Mild slices, with an F1-score of 0.8608, and reasonable performance for Severe slices, with an F1-score of 0.6877. Moderate was the most challenging class, achieving an F1-score of 0.2979. This finding is consistent with Zeng et al.~\cite{zeng2025liver}, which also identified intermediate-stage classification as the most challenging setting.

\begin{table}[!htbp]
\centering
\caption{Class-wise slice-level performance of CirrGuide.}
\label{tab:classwise_results}
\setlength{\tabcolsep}{6pt}
\begin{tabular}{lcccc}
\hline
\textbf{Class} & \textbf{Precision} & \textbf{Recall} & \textbf{F1-score} & \textbf{Support} \\
\hline
Mild & 0.8646 & 0.8571 & 0.8608 & 350 \\
Moderate & 0.2819 & 0.3158 & 0.2979 & 133 \\
Severe & 0.7143 & 0.6630 & 0.6877 & 181 \\
\hline
\end{tabular}
\end{table}

\textbf{Segmentation Guidance Ablation:} 
Table~\ref{tab:mask_ablation} evaluates the contribution of segmentation
guidance to severity classification. Attention without mask guidance achieved
a macro F1-score of 47.84\%, compared with 61.55\% for CirrGuide, supporting
the importance of anatomical guidance beyond generic attention.

\begin{table}[!htbp]
\centering
\caption{Ablation of segmentation guidance on the CirrMRI600+ T2W test set.
QWK denotes quadratic weighted kappa and MAE denotes ordinal mean absolute error.}
\label{tab:mask_ablation}
\setlength{\tabcolsep}{5pt}
\begin{tabular}{lcccc}
\hline
\textbf{Configuration} &
\textbf{Acc. $\uparrow$} &
\textbf{Macro-F1 $\uparrow$} &
\textbf{QWK $\uparrow$} &
\textbf{MAE $\downarrow$} \\
\hline
Global Features Only
& 0.5828 & 0.5535 & 0.6946 & 0.4187 \\

No Mask + Attention
& 0.5587 & 0.4784 & 0.6323 & 0.4849 \\

Hard Mask + Attention
& 0.6039 & 0.5608 & 0.7096 & 0.4006 \\

Soft Mask + Average Pooling
& 0.5783 & 0.5401 & 0.6489 & 0.4367 \\

\textbf{CirrGuide}
& \textbf{0.6958}
& \textbf{0.6155}
& \textbf{0.7758}
& \textbf{0.3102} \\
\hline
\end{tabular}
\end{table}

Hard predicted-mask guidance improved macro F1 to 56.08\%, while replacing learned
attention with soft-mask-weighted average pooling achieved 54.01\%. CirrGuide,
which combines soft predicted-mask guidance with learned regional attention,
achieved the strongest overall performance, with 69.58\% accuracy, 61.55\%
macro F1, a QWK of 0.7758, and an ordinal MAE of 0.3102.

\textbf{Patient-level Analysis:} Patient-level performance is obtained by averaging slice-level softmax probabilities across all slices of each test case and assigning the class with the highest mean probability. Using this probability-aggregation strategy, CirrGuide achieved a patient-level accuracy of 0.7097 and a macro F1-score of 0.5859 across the 31 official T2W test cases, correctly classifying 15/16 Mild and 6/9 Severe cases but only 1/6 Moderate cases, highlighting the difficulty of intermediate-stage classification. We further investigated learned patient-level aggregation using frozen CirrGuide slice representations. In this controlled analysis, attention pooling achieved a macro F1-score of 0.5715, whereas BiGRU-based inter-slice modeling reduced it to 0.5316, indicating no benefit from additional recurrent sequence modeling. A weighted CORAL-style ordinal objective achieved a QWK of 0.7896 and an ordinal MAE of 0.2903, but a macro F1-score of 0.5481. These results suggest that lightweight attention aggregation was preferable to recurrent sequence modeling, while ordinal supervision mainly improved severity-aware agreement rather than balanced three-class discrimination.

\begin{figure}[!htbp]
\centering
\includegraphics[width=\textwidth]{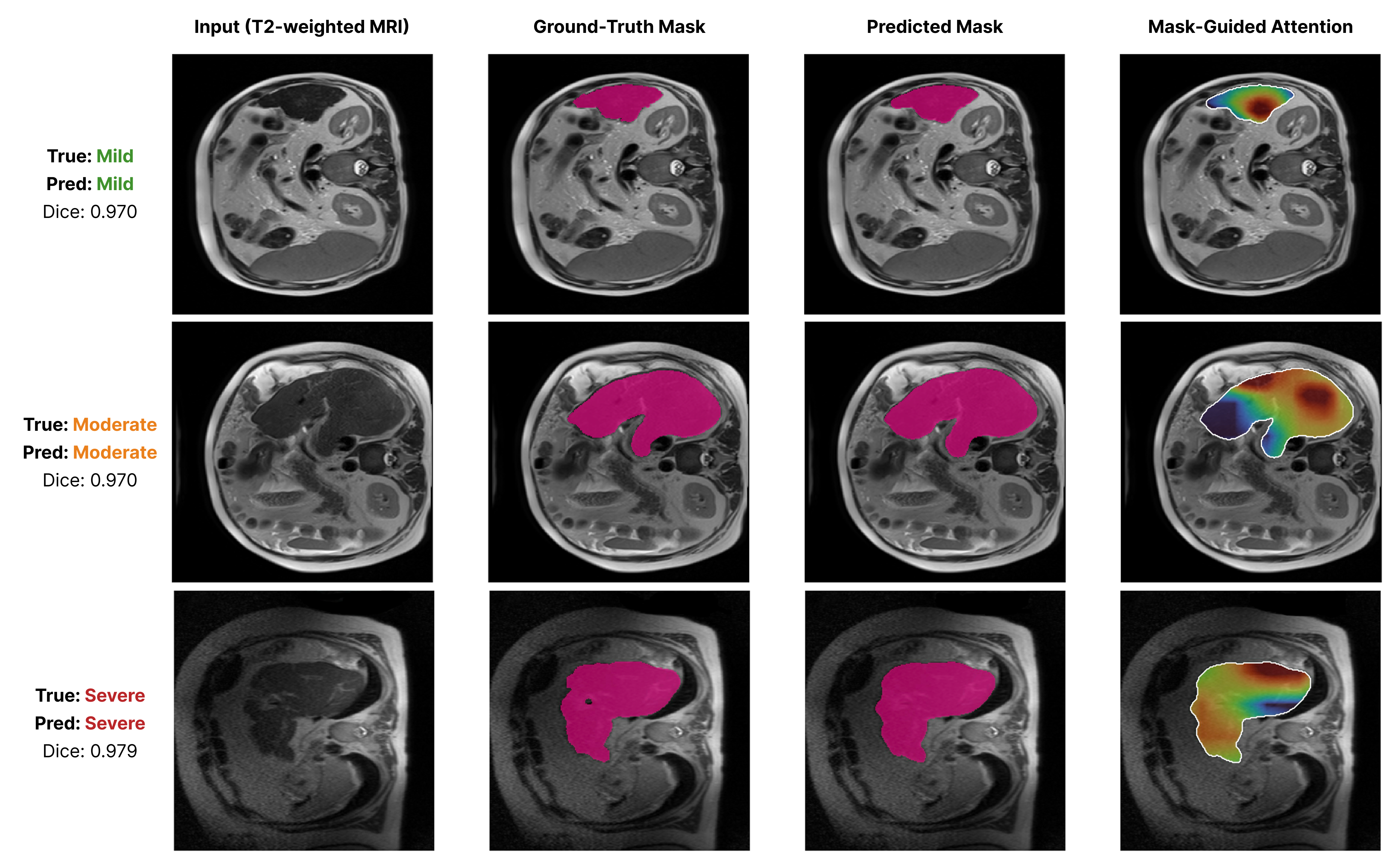}
\caption{Qualitative examples of CirrGuide predictions. Pink overlays show ground-truth and predicted cirrhotic liver masks. Warmer colors in the attention map indicate higher mask-guided pooling attention within the predicted segmentation region.}
\label{fig:qualitative_results}
\end{figure}

\textbf{Qualitative Analysis:} 
Figure~\ref{fig:qualitative_results} presents
representative Mild, Moderate, and Severe cases. 

The predicted masks show strong
agreement with the ground-truth cirrhotic liver regions across severity levels,
supporting the quantitative segmentation results. The attention maps further show that
the classifier does not weight the entire segmented liver uniformly; instead, it assigns
higher attention to selected regions within the predicted mask. This behavior aligns
with the design of CirrGuide, where segmentation provides anatomical localization and
mask-guided attention pooling encourages the classifier to focus on discriminative liver
subregions.

\section{Conclusion}

We proposed CirrGuide, a cascaded segmentation-guided framework for cirrhotic liver
region segmentation and severity classification from T2W MRI. CirrGuide first
predicts a soft cirrhotic liver mask and then uses it as an anatomical prior for
mask-guided attention pooling, explicitly linking localization with severity prediction.
Experiments on the official CirrMRI600+ T2W 2D split show that CirrGuide improves both
segmentation and classification performance compared with segmentation-only,
classification-only, and multi-task baselines. These results demonstrate the value
of predicted segmentation masks as anatomical guidance for T2W MRI-based cirrhosis
severity classification.

\begin{credits}

\subsubsection{\discintname}
The authors declare that they have no competing interests relevant to the content of this article.

\end{credits}

\bibliographystyle{splncs04}
\bibliography{ref}

\end{document}